\documentclass[sigconf]{acmart}

\usepackage{booktabs} 
\usepackage{blindtext, graphicx}
\usepackage{amssymb}
\usepackage[]{algorithm2e}
\usepackage{mathtools}
\usepackage{amsmath}
\usepackage{fixltx2e}
\usepackage{algpseudocode}
\usepackage[]{algorithm2e}
\usepackage{gensymb}
\usepackage{placeins}
\usepackage{graphicx}
\usepackage{tabularx}
\usepackage{setspace}
\usepackage{hyperref}
\usepackage{float}
\usepackage{multirow}
\usepackage{booktabs}
\usepackage{epstopdf} 
\usepackage{subcaption}
\graphicspath{{Images/}}

\setcopyright{rightsretained}

\acmDOI{10.475/123_4}

\acmISBN{123-4567-24-567/08/06}

\copyrightyear{2017}
\acmYear{2017}
\setcopyright{acmcopyright}
\acmConference{GECCO '17 Companion}{July 15-19, 2017}{Berlin, Germany}\acmPrice{15.00}\acmDOI{http://dx.doi.org/10.1145/3067695.3082511}
\acmISBN{978-1-4503-4939-0/17/07}

\begin{document}
\title{Exploiting Active Subspaces in Global Optimization: How Complex is your Problem?}

\author{Pramudita Satria Palar}

\affiliation{%
  \institution{Institute of Fluid Science, Tohoku University}
  \streetaddress{2 Chome-1-1 Katahira}
  \city{Sendai} 
  \state{Japan} 
  \postcode{980-8577}
}
\email{pramudita.satria.palar.a5@tohoku.ac.jp}

\author{Koji Shimoyama}

\affiliation{%
  \institution{Institute of Fluid Science, Tohoku University}
  \streetaddress{2 Chome-1-1 Katahira}
  \city{Sendai} 
  \state{Japan} 
  \postcode{980-8577}
}
\email{shimoyama@tohoku.ac.jp}

\renewcommand{\shortauthors}{Pramudita Satria Palar and Koji Shimoyama}

\begin{abstract}
When applying optimization method to solve real-world problems, the possession of prior knowledge and preliminary analysis on the landscape of global optimization problems can give us an insight into the complexity of the problem. This knowledge can better inform us in deciding what optimization method should be used to tackle the problem. However, this analysis becomes problematic when the dimensionality of the problem is high. This paper presents a framework to take a deeper look on the global optimization problem to be tackled: by analyzing the low-dimensional representations of the problem through discovering the active subspaces of the given problem. The virtue of this is that the problem's complexity can be visualized in a one or two-dimensional plot, thus allow one to get a better grip about the problem's difficulty. One could then have a better idea regarding the complexity of their problem to determine the choice of global optimizer or what surrogate-model type to be used. Furthermore, we also demonstrate how the active subspaces can be used to perform design exploration and analysis.  
\end{abstract}

%
%


\maketitle

\section{Introduction}
Since the dawn of evolutionary computation, metaheuristics methods have been widely studied and applied to solve many real-world optimization problems. Among several advantages of metaheuristic optimization methods that made them attractive are their higher likelihood to discover the global optimum, gradient-free nature, and the capability to discover the Pareto front in a single run (for the multi-objective case). The most widely used metaheuristic is arguably evolutionary algorithm (EA) that relies on the principle of natural evolution to guide the discovery of the optimum solution. In spite of these advantages, the main drawback of metaheuristic methods is still their expensive cost. Typical metaheuristic optimizers might call more than a thousand simulations to locate the optimum solution(s) or location near the true optimum(s). In cases with expensive function evaluation such as computer-aided engineering design, direct use of metaheuristics is strictly prohibitive. Although metaheuristic can still be applied to solve such cases, the found optimized solution(s) might still be far from the true optimum. The surrogate model came as a remedy to handle this issue of computationally expensive function evaluation. The surrogate model works by aiding the optimizer through a cheap approximation of the true function. Regardless of the use of surrogate or non-surrogate-based optimizer, it is common that one optimizes the problem without any prior knowledge regarding the problem being investigated. The possession of this prior knowledge might help one to better decide the optimization strategy to be applied.  

In the field of engineering optimization, there were some attempts to analyze the behavior and complexity of the problem before or during the optimization. The information gained from this analysis could be used to reduce the complexity of the problem or to gain more understanding about the problem being solved. Jeong et al. applied ANalysis Of VAriance (ANOVA) to analyze the contribution of each individual output to the objective functions via a Kriging surrogate model~\cite{jeong2005efficient}. ANOVA also allows one to reduce the number of design variables to reduce the complexity of the problem beforehand. However, ANOVA cannot give a clue about the complete picture of the problem being solved. For example, it is difficult to infer the information about the non-linearity or the multi-modality behavior of any function by using ANOVA. The use of ANOVA comes with a warning where optimum design might not be achieved by merely reducing the number of design variables. In this regard, Ghisu et al. enhance the multi-objective Tabu search with principal component analysis to transform the original coordinate into a more optimum representation of the design space~\cite{ghisu2008accelerating} which does not need to be aligned with the original coordinate. More recently, Deb applied the high-dimensional model representation (HDMR) framework to narrowing the search space of the problem~\cite{roy2016high}.

In solving real-world global optimization problems, it is important to properly select the type of optimizer to be employed. Local optimizers such as sequential quadratic programming is effective to solve global optimization problems with only one global optimum. However, local optimizer is not suitable to solve multi-modal problems, instead, multi-start local search should be utilized. On the other hand, global optimizers such as metaheuristics methods has a higher likelihood to discover the true optimum of the multimodal problem at the expense of computational cost. The right choice of the optimizer mainly depends on the complexity of the problem itself. It is relatively easy to analyze the problem's complexity in one or two-dimensional problem since it can be plotted in an informative visual way. This kind of analysis can also aid the decision process of what type of surrogate model to be used to solve the problem. However, it is worth noting that such analysis becomes problematic in high-dimension since it is not possible to visualize the problem's complexity. 

One way to perform this visual analysis is to discover the possible low-dimensional structure of the problem, where the information can be embedded into a visualizable lower-dimensional subspace. Based on the experience and shreds of evidence from experts, there are many cases where the behavior of the real-world problem (either optimization, uncertainty quantification, or sensitivity analysis) is not as difficult as the test problems widely considered in metaheuristic-based optimization literature. There is a possibility that the problem's landscape only exhibits a few important dimensions while the rest of them are relatively unimportant, rendering the possibility of creating an informative one or two-dimensional plot. This kind of plot is even informative in problems that cannot be compressed to low-dimensional representation. By analyzing the plot in such cases, one can infer the highly complex structure of the problem that needs a more special treatment or method to solve the given optimization problem. Some methods that can be used to perform such tasks are active subspace method (ASM)~\cite{constantine2014active} and sliced inverse regression (SIR)~\cite{li1991sliced, li2016inverse}. In particular, the ASM has been employed to aid the process of uncertainty quantification,~\cite{constantine2015exploiting} sensitivity analysis~\cite{jefferson2016reprint}, and aerodynamic design optimization in reduced space~\cite{lukaczyk2014active, othmer2016active}. The ASM is particularly attractive since it is relevant to the goal of optimization in computationally expensive problems in order to solve the problem efficiently under limited computational budget. In this paper, we explore the capability of the ASM to perform exploratory analysis of the design space instead of executing the optimization in the reduced space, which is still an area of active research. In this sense, we use the ASM to better inform us about the complexity level of a global optimization problem. 

Our objective in this paper is to introduce and investigate the usefulness of ASM as a tool to analyze the complexity of the global optimization problem by discovering the possible low-dimensional representation of the problem. We demonstrate the framework on three synthetic and two real-world problems, in which the application of the ASM reveals some important structures and features that helped us in understanding the problem's complexity.

\section{Low-dimensional representation of global optimization problem}
In this paper, we have the interest to solve optimization problems with the gradient-free method. This can be performed by using metaheuristics or surrogate-based optimization method. In many occasions, one directly solves the problem with any optimizer in possession without making a guess about the complexity of the problem. This, in turn, can result in the inefficiency on the problem solving itself. For example, using a metaheuristic optimizer for the optimization of a unimodal problem is quite an overkill. On the other hand, using a one-shot local optimizer for multi-modal problems results in a huge risk of missing the global optimum. Making such inference is not difficult for one and two-dimensional problem but it becomes troublesome in a higher-dimensional problem. Before one performs global optimization, it is useful to analyze and infer the complexity of the problem beforehand. This information would be useful when one wants to decide which type of optimizer or surrogate model to be used. For example, one can predict whether the problem is unimodal or multimodal by using this prior information. If the problem is an unimodal one, it is better to use simple surrogate-based optimizer or a local search method than the expensive metaheuristic based optimization. Another example is to detect the smoothness of the problem's landscape to investigate whether the problem is really smooth or exhibits discontinuity to some degree. Applying polynomial-based surrogate model in a discontinuous problem is obsolete since polynomial cannot properly capture discontinuity in the response surface, hence the use of non-parametric surrogate models such as Kriging might be more helpful. This kind of diagnosis can be performed by finding a low-dimensional representation of the problem that allows us to take a peek on the function's complexity. In this regard, the low-dimensional representation of the input-output relationship is what we seek.

The key concept to this analysis lies on the concept of low-dimensional representation, where the problem's complexity can be easier to analyze/solve if the problems are transformed into a low-dimensional subspace that could better explain the problem's variability.  We advocate the use of ASM to perform such task. One objective of this paper is to further introduce this methodology to the wider community of optimization, metaheuristics, and also practitioners alike.

\subsection{Active subspace method}
The low-dimensional representation of a problem can also be expressed in terms of sufficient dimension. Sufficient dimension is the subspace of a problem that explains most of the variability of the function. ASM uses the outer product of the gradient information to discover this low-dimensional representation expressed in terms of eigenpairs. The sufficient dimension itself does not need to be aligned with the original coordinate system since the most active direction might not lie in the original untransformed coordinate. We refer to~\cite{constantine2014active} for the following explanation of ASM. 

The first necessary step of the ASM is to compute the averaged outer product of the objective function gradient $\nabla f$, denoted as $\textrm{\textbf{C}}$. This can be computed by averaging over $M$ samples as follows:

\begin{equation}
\label{eq:ASM1}
\textrm{\textbf{C}} \approx \frac{1}{M}\Sigma_{i=1}^{M}\nabla f(\boldsymbol{x}^{(i)}) f(\boldsymbol{x}^{(i)})^{T}.
\end{equation}

After $\textrm{\textbf{C}}$ is obtained, the next step is to perform the eigendecomposition of $\textrm{\textbf{C}}$, reads as
\begin{equation}
\label{eq:ASM2}
\textbf{C} = \textbf{W} \Lambda \textbf{W},
\end{equation}
where 
\begin{equation}
\label{eq:ASM3}
\textbf{W} = [\textbf{w}_{1},\ldots,\textbf{w}_{m} ],~ \Lambda=\textrm{diag}(\lambda_{1},\ldots,\lambda_{m})
\end{equation}
with $\textbf{w}_{j}$ is the eigenvectors which are sorted according to the descending eigenvalues $\lambda_{j}$, so $\lambda_{1}\geq \ldots,\lambda_{m} $, where $m$ is the dimensionality of the problem. Reducing the dimensionality of the problem can then be performed by taking the first $n$ eigenvectors to construct the basis of the $n-$dimensional active subspace $\textbf{W}_{1}$ as a partition of $\textbf{W}$, reads as
\begin{equation}
\textbf{W} = [\textbf{W}_{1},\textbf{W}_{2}],~\Lambda=
\begin{bmatrix}
\Lambda_{1} & \\
 & \Lambda_{2}
\end{bmatrix},
\end{equation}
where $\Lambda_{1}$ contains the first $n$ eigenvectors, with $n<m$ and $\textbf{W}_{1}$ is of $m\times n $ size. The design variables can now be projected onto the active subspace $span(\textbf{U})$ to obtain the rotated coordinates $\boldsymbol{x}_{r}$ defined as:
\begin{equation}
\boldsymbol{x}_{r} = \textbf{W}_{1}^{T}\boldsymbol{x}, \boldsymbol{x}_{r}\in \mathbb{R}^{n}.
\end{equation}

The objective function can then be expressed in the rotated coordinate as
\begin{equation}
	f(\boldsymbol{x}) \approx g(\textbf{W}_{1}^{T}\boldsymbol{x}).
\end{equation}

As one can see from this formulation, the dimensionality $n$ can be set to either one or two which makes visualization (become) possible. The plot might indicate the existence of sufficient dimension in the problem being tackled if the first and second eigenvalues are significantly large compared to the others. However, a complex behavior in the one/two-dimensional plot reduced coordinate might indicate that the problem is multi-modal and highly non-linear. 

Another benefit of evaluating the active subspace is that we can compute the contribution of each variable with the global sensitivity metric derived from the active-subspace called the activity scores~\cite{constantine2015global}. The activity score $\alpha$ for variable $i$ is computed as follows:
\begin{equation}
	\alpha_{i} = \alpha_{i}(n) = \sum_{j=1}^{n}\lambda_{j}w_{i,j}^{2},~~i=1,\ldots,m.
\end{equation}

This allows us to rank which variables are the most and least important, which could give us a further insight relating to the true complexity of the problem and the physical insight itself. When $n=m$, the activity scores become the derivative-based global sensitivity metric~\cite{constantine2015global}. In this paper, we use the derivative-based global sensitivity metric by setting $n$ equals to $m$.

\subsection{Estimating the Active Subspace with Surrogate model}
The original active-subspace method needs the gradient information of the function to find the underlying active subspace. Although in some cases such as computational fluid dynamics (CFD)-based design this gradient can be computed via adjoint-method, gradient computation remains a bottleneck in many engineering problems. To cope with this problem, one can build a surrogate model of the function first and then estimate the gradient information using this surrogate model. This strategy has been investigated in the context of car aerodynamics in order to discover the existence of the active subspace~\cite{othmer2016active}.  Since surrogate model can be evaluated cheaply, using finite-difference is an effective way to estimate the gradient based on the surrogate model.

To limit our scope of discussion in this paper, we only used two types of surrogate models to perform the ASM: Kriging~\cite{krige1951statistical} and sparse polynomial chaos expansion (PCE)~\cite{blatman2011adaptive}. Kriging is a non-parametric surrogate model that can capture non-linear trend in the function due to its flexibility. The drawback of Kriging is that its training time can be very long especially when the dimensionality and the number of samples are high. On the other hand, sparse PCE is very fast but it cannot capture a highly non-linear surface properly. Kriging is suitable when the number of initial samples and dimensionality is low since Kriging training time is still reasonable in this range. However, Kriging training time can be burdensome when the number of initial samples and dimensionality is high, say $k>200$ and $m>15$. Sparse PCE which works by automatically detecting the important polynomial terms is more suitable for high-dimensionality problems due to its fast training time.

Brief explanation of each surrogate model is explained below:
\subsubsection{Kriging}
Kriging approximates the true function with a combination of the basis functions of
\begin{equation}\label{eq:2.5}
\psi^{(i)}=\text{exp}\bigg(-\sum_{j=1}^{m}\theta_{j}|x_{j}^{(i)}-x_{j}|^{p_{j}}\bigg).
\end{equation}
The basis of Kriging model is a vector $\boldsymbol{\theta}=\{\theta_1,\theta_2,...,\theta_{m}\}^{\text{T}}$. The exponent $\boldsymbol{p}=\{p_{1},p_{2},...,p_{m} \}^{\text{T}}$ is also tunable but we set a fix value of $p=2$ for simplicity purpose. Here,  $\boldsymbol{\theta}$ are optimized by maximizing the likelihood function. After the optimum hyperparameters were found, the Kriging predictor reads as
\begin{equation}\label{eq:2.11}
\hat{f}_{KRG}(\boldsymbol{x})=\hat{\mu}_{K}+\boldsymbol{\psi}^{\text{T}}\boldsymbol{\Psi}^{-1}(y-\boldsymbol{1}\hat{\mu}_{K}),
\end{equation}
where $\hat{\mu}_{K}$, $\boldsymbol{\psi}$, and $\boldsymbol{\Psi}$ are the mean of the Kriging approximation, correlation matrix between the experimental design and $\boldsymbol{x}$, and the correlation matrix between all experimental design, respectively. More detailed implementation of Kriging method can be found elsewhere (see ~\cite{j2008kriging} for example).

\subsubsection{Sparse polynomial chaos expansion}
PCE approximates the function with the sum of orthogonal polynomials $\boldsymbol{\Theta} = \{\Theta_{0},\ldots,\Theta_{P}\}$. For optimization purpose, Legendre polynomials are used due to the bounded nature of the optimization problem.

PCE works by approximating $f(\boldsymbol{x})$ with
\begin{equation}
	\hat{f}_{PC}(\boldsymbol{x}) = \sum_{i=0}^{P}\alpha_{i}\Theta_{i}(\boldsymbol{x}).
\end{equation}
To find the optimum set of polynomial bases and compute the coefficients, the sparse PCE representation employs least-angle-regression (LARS)~\cite{blatman2011adaptive}.

The gradient information for the ASM can be simply obtained analytically or by using finite difference which is now very cheap since it is computed using the already built surrogate model. The pseudocode to find the active subspace and plotting in reduced coordinate is detailed in Algorithm~\ref{alg:ASM}.
\begin{algorithm}
	Prepare the initial experimental design $\boldsymbol{\mathcal{X}}$;\\
	Evaluate the output $\boldsymbol{y}$;\\
	Build surrogate model $\hat{f}(\boldsymbol{x})$ using $\boldsymbol{\mathcal{X}}$ and $\boldsymbol{y}$;\\
	Estimate the gradient information $\nabla f(\boldsymbol{x})$ for each design in $\boldsymbol{\mathcal{X}}$ using $\hat{f}(\boldsymbol{x})$ (i.e. finite difference) ;\\
	Obtain the eigenvectors $\textbf{W}$ and eigenvalues $\Lambda$ using Eqs.\ref{eq:ASM1} and \ref{eq:ASM2};\\
	Obtain the partitioned $\textbf{W}_{1}$ and $\Lambda_{1}$ with $n=1$ or $n=2$ (Eq.\ref{eq:ASM3});\\
	Obtain the rotated coordinate $\boldsymbol{x}_{r} \in \mathbb{R}^{n}$ and plot the $\boldsymbol{x}_{r}$ versus $\boldsymbol{y}$;\\
	\caption{Pseudocode of design space visualization using ASM.}
	\label{alg:ASM}
\end{algorithm}

The take home point is that one should at least take a peek on the decision variable-objective function relationship by using the ASM. We will demonstrate the usefulness of this visualization framework, and the ASM in general, in the next section. 

\section{Computational demonstration}
In this section, we demonstrate the various usefulness of ASM on some functions in order to perform preliminary exploratory analysis of the global optimization problem's landscape. The first three problems are algebraic, while the last two problems are real-world problems which were evaluated using partial differential equation solver.

\subsection{Example 1: Zakharov function}
The demonstration was firstly performed on the Zakharov function. Here, we want to demonstrate that the ASM has the capability to visualize the existence of the single global optimum in the Zakharov function. The Zakharov function is expressed as:
\begin{equation}
f(\boldsymbol{x}) = \sum_{i=1}^{m}x_{i}^{2}+\bigg(\sum_{i=1}^{m}0.5ix_{i}\bigg)^{2}+\bigg(\sum_{i=1}^{m}0.5ix_{i}\bigg)^{4}
\end{equation}
where the function is evaluated on the hypercube $x_{i}\in[-5,10]$ for all $i=1,\ldots,m$. The dimensionality and the number of initial samples for this problem was set to 20 and 200, respectively. 

\begin{figure}
		\centering
			\includegraphics[width=0.7\columnwidth]{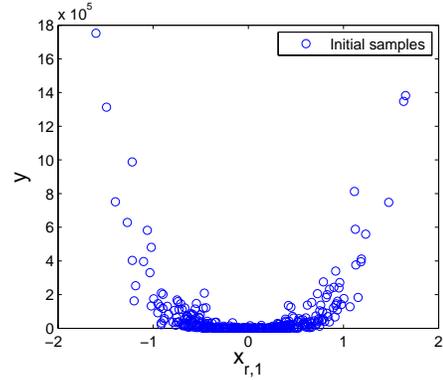}%
			\caption{Reduced coordinate versus $y$ on Zakharov function.}%
			\label{fig:ACS_ZAKHAROV}%
	\end{figure}

Since the implementation of Kriging for this problem would be expensive, we have to rely on PCE to create the surrogate model for the Zakharov function. We then estimate the active subspace based on the PCE model where the result is shown in Fig.\ref{fig:ACS_ZAKHAROV}. Upon rotating the coordinate based on the gradient information from the PCE surrogate model, we are now able to detect the presence of single global optimum on the Zakharov function. This information is difficult to observe if we directly plot one original variable versus the output information. We can then make use of this plot to decide which optimizer that we should use. For example, the presence of single global optimum suggests us to use a simple local search optimization method from the current optimum point to perform optimization instead of applying expensive metaheuristic technique. 

\subsection{Example 2: Hartman-6 function}
\begin{figure}
		\centering
			\includegraphics[width=0.7\columnwidth]{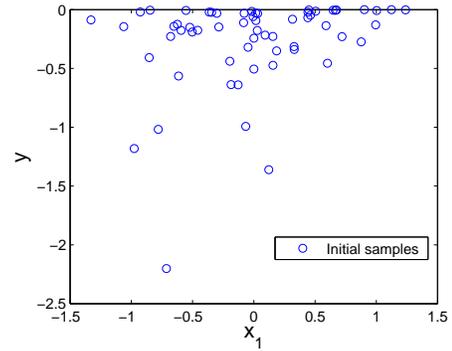}%
			\caption{Reduced coordinate versus $y$ on Hartman-6 problem.}%
			\label{fig:ACS_HART6}%
\end{figure}

The Hartman-6 function is expressed as follows: 
\begin{equation}
		y(\boldsymbol{x}) = -\sum_{i=1}^{4}c_{i}\text{exp } \bigg\{-\sum_{j=1}^{m}A_{ij}(x_{j}-P_{ij})^2\bigg\}
\end{equation}
where $\boldsymbol{x}=(x_{1},x_{2},\ldots,x_{m})^{T}, x_{i}\in[0,1]$. Details of $\textbf{A}$, $\textbf{P}$, and $\boldsymbol{c}$ can be found in~\cite{dixon1978global}. The dimensionality of this problem is 6 with the initial sample size was set to 60. Kriging surrogate model was employed since the training time is relatively fast for this problem. The one-dimensional plot in the reduced coordinate is then shown in Fig.~\ref{fig:ACS_HART6}. In contrary to the previous problem, the plot in the reduced coordinate does not show any evidence about the existence of the sufficient dimension. However, the plot indicates that the landscape of the function is probably multi-modal due to the existence of this complex behavior (indeed, the Hartman-6 function has 6 local minima). Although not shown here, the two-dimensional plot also reveals similar behavior. Another knowledge that can be inferred from this plot is the possibility of a highly non-linear behavior of the function which suggests the use of non-parametric surrogate models instead of the parametric one (if one wants to employ surrogate model). If no surrogate model is used, it is suggested to use a metaheuristics global optimizer or a multi-start local search procedure to ensure that the optimum of this problem is found. A simple one-shot local optimization should be avoided since there is a high possibility of missing the true optimum. 

\subsection{Example 3: Four bar trusses problem}
The third problem is still algebraic but more realistic test problem. The problem is the four bar trusses which was first introduced by~\cite{stadler1993multicriteria} and has been widely used as a test problem for multi-objective optimization and structural optimization. The dimensionality and the initial sample size for this problem were 4 and 40, respectively. The four bar problem is defined as follows:
\begin{multline}
	f_{1}(\boldsymbol{x}) = L(2x_{1}+\sqrt{2}x_{2}+\sqrt{x_{3}}+x_{4}) \\
	f_{2}(\boldsymbol{x}) = \frac{FL}{E}(\frac{2}{x_{1}}+\frac{2\sqrt{2}}{x_{2}}-\frac{2\sqrt{2}}{x_{3}}+\frac{2}{x_{4}})
\end{multline}

where $(F/\sigma)\leq x_{1} \leq 3(F/\sigma)$, $\sqrt{2}(F/\sigma)\leq x_{2} \leq 3(F/\sigma)$, $\sqrt{2}(F/\sigma)\leq x_{3} \leq 3(F/\sigma)$, and $(F/\sigma)\leq x_{4} \leq 3(F/\sigma)$. Here, $F=10$kN, $E=2\times10^{5}$kN/cm$^{2}$, $L=200$cm, $\sigma=10$kN/cm$^{2}$. 

\begin{figure}
		\centering
		
		\begin{subfigure}{0.7\columnwidth}
			\includegraphics[width=1\columnwidth]{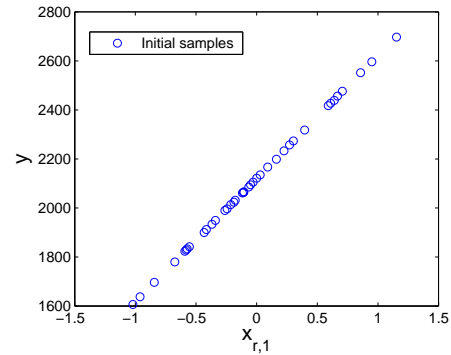}%
			\caption{Reduced coordinate versus $y_{1}$.}%
			\label{ACS_fourbar_1.eps}
		\end{subfigure}\hfill%
		\begin{subfigure}{.7\columnwidth}
			\includegraphics[width=1\columnwidth]{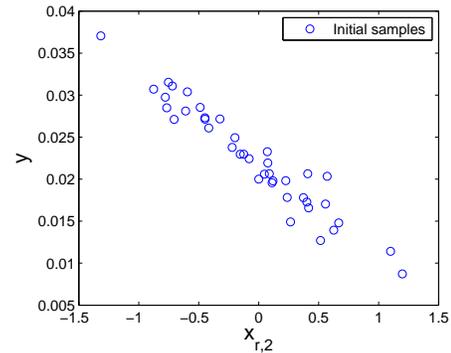}%
			\caption{Reduced coordinate versus $y_{2}$.}%
			\label{fig:ACS_fourbar_2}%
		\end{subfigure}\hfill%
		\caption{One-dimensional reduced coordinate plot on the fourbar problem.}
		\label{fig:ACS_fourbar}
	\end{figure}

As we can see from the depiction in Fig.~\ref{fig:ACS_fourbar}, both objectives can be sufficiently approximated by a one-dimensional active subspace that explains most of the variability of the function. We can infer from this result that the problem has a high chance of being a unimodal problem if single-objective optimization is performed for each objective (Although it is obvious that the first objective has a linear expression, this kind of observation is particularly useful if the function is a black-box one). For problems like this, using metaheuristic optimizer might be an overkill since the problem is a unimodal function, especially if the function evaluation is not cheap. It is then better to employ local search or gradient-based optimizer (if the gradient is available) for single-objective optimization. When one wants to employs a surrogate to tackle this problem, a 1st or 2nd order polynomial might be more suitable than RBF due to the linear or almost linear behavior of the function. However, it won't hurt to apply EA to solve this problem when the function evaluation is cheap since one now has a higher confidence of discovering the optimum solution due to the unimodality of the function. 

\begin{figure}
			\centering
				\includegraphics[width=0.7\columnwidth]{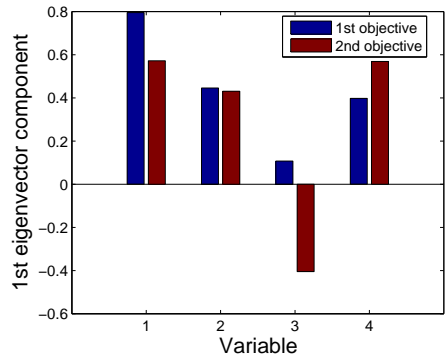}%
				\caption{Components of the 1st eigenvector on the fourbar problem.}%
				\label{fig:ACS_fourbar_BAR}%
			\end{figure}

When concerning multi-objective optimization, Figs.~\ref{fig:ACS_fourbar} and ~\ref{fig:ACS_fourbar_BAR} show some useful information regarding the difficulty of the multi-objective four bar problem. Firstly, Fig.~\ref{fig:ACS_fourbar_BAR} depicts the component of the first eigenvectors of the plot shown in Fig.~\ref{fig:ACS_fourbar}. Basically, this figure tells us the relationship between the objective and the variables variation. As for example, a high value of the first objective can be achieved by increasing all variables in the first eigenvector direction, with variables 1 and 3 have the highest and the lowest contribution. The sign here tells us whether we should increase or decrease the value of the variables to achieve minimum or maximum value of the objective. Furthermore, we can see that the first eigenvectors components of both objectives have similar tendency but mainly differ on the direction of the third variable. Based on this plot, we can further make an inference that the two objectives are indeed conflicting to each other, since decreasing the first objective will increase the second objective.

\subsection{Example 4: Viscous Transonic Airfoil Design Preliminary Analysis}
\begin{figure}
			\centering
				\includegraphics[width=0.9\columnwidth]{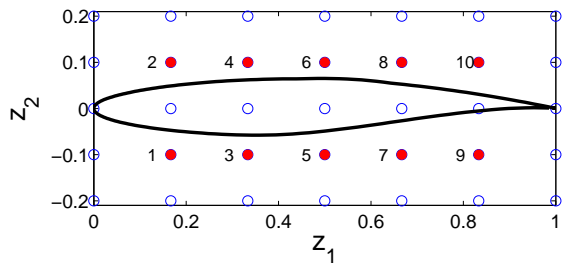}%
				\caption{FFD box used in the viscous transonic airfoil design problem.}%
				\label{fig:ACTIVESUB_VISCOUS_AIRFOIL}%

		\end{figure}

\begin{figure}
			\centering
			
			\begin{subfigure}{0.7\columnwidth}
				\includegraphics[width=1\columnwidth]{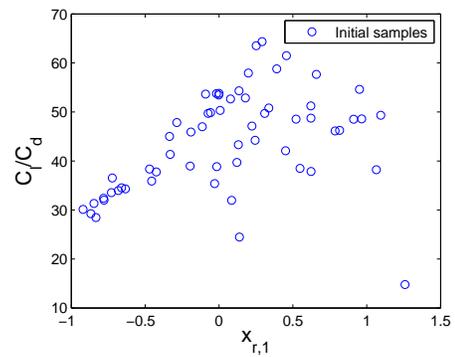}%
				\caption{One-dimensional plot.}%
				\label{fig:ACTIVESUB_VISCOUS_1}%
			\end{subfigure}\hfill%
			\begin{subfigure}{.7\columnwidth}
				\includegraphics[width=1\columnwidth]{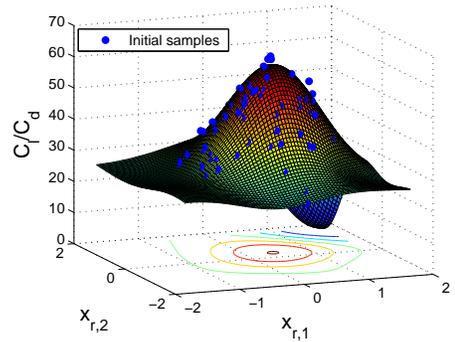}%
				\caption{Two-dimensional plot.}%
				\label{fig:ACTIVESUB_VISCOUS_2}%
			\end{subfigure}\hfill%
			\caption{Reduced coordinate plot on the viscous transonic airfoil problem.}
			\label{fig:ACTIVESUB_VISCOUS}
		\end{figure}

In this case, we wanted to investigate which type of surrogate is better to approximate the given aerodynamic function. The problem is the viscous transonic airfoil redesign of RAE 2822 airfoil in the flight condition of Mach number ($M$)=0.729 and angle of attack $(AoA)=2.31^{0}$, with the maximization of lift to drag ratio ($C_{l}/C_{d}$) as the objective function. A Reynolds-averaged Navier-Stokes (RANS) CFD code was employed to solve this problem. To alter the geometry of the airfoil, free-form-deformation (FFD) technique was applied where the depiction of the airfoil's geometry and the FFD box are shown in Fig.~\ref{fig:ACTIVESUB_VISCOUS_AIRFOIL}. Only ten FFD points were allowed to move, and the upper bound and lower bound of the movement in $z_{2}$ direction are 0.05 and -0.05, respectively (the movement in $z_{1}$ direction was locked). The dimensionality and the initial sample size for this problem are 10 and 70, respectively. Kriging surrogate model was employed to discover the active subspaces in this problem.
The plot in the one-dimensional active subspace does not reveal enough information for us to infer since there is no clear trend observed (see Fig.~\ref{fig:ACTIVESUB_VISCOUS_1}). The two-dimensional plot in Fig.~\ref{fig:ACTIVESUB_VISCOUS_2}, however, display a clear and interesting trend. Here, we approximated the response surface in the two-dimensional reduced coordinate with Kriging regression to better depict the underlying trend. The first eigenvector alone explains 58.62\% of the total variance, while both the first and second eigenvector describes 88.98\%. From the two-dimensional plot, we can see that there exists a non-linear trend in the response surface whose shape is like a combination of radial basis functions. Indeed, applying Kriging and sparse PCE results in the mean absolute error of 4.5295 and 7.5023, respectively, when a test was performed using 30 independent validation samples. In this regard, the plot in the reduced coordinate had informed us to utilize Kriging with RBF kernel rather than the sparse PCE (besides the information that comes from cross-validation).

\begin{figure}
			\centering
				\includegraphics[width=0.7\columnwidth]{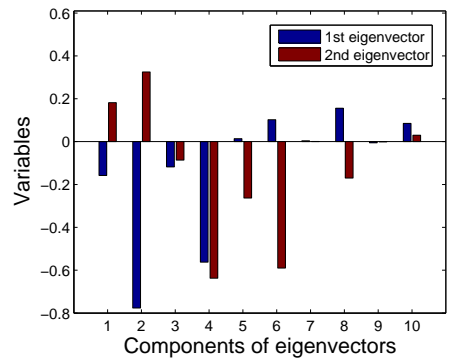}%
				\caption{Components of the 1st and 2nd eigenvectors on the viscous transonic airfoil problem. }%
				\label{fig:ACTIVESUB_VISCOUS_BAR}%

		\end{figure}

As an additional information provided by the active subspace, Fig.~\ref{fig:ACTIVESUB_VISCOUS_BAR} shows the component of the 1st and 2nd eigenvectors. The barplot mainly tells us that the upper FFD points have the largest contribution to the objective function variance. The derivative-based global sensitivity metrics shown in Table~\ref{tbl:ViscousActivity} further reveal that variables 2 and 4, which are located near the leading edge of the airfoil, are the largest contributors of all. We can also see that the contribution of variable 7 and 9 are so small that they can be safely neglected. 

	\begin{table*}
				\centering
				\begin{tabular}{ccccccccccc} \hline 
					No. & 1 & 2 & 3 & 4 & 5 & 6 & 7 & 8 & 9 & 10 \\ \hline
					Score & 0.14 & 1.71 & 0.09 & 1.38 & 0.16 & 0.55 & 0.00 & 0.24 & 0.00 & 0.05 \\
\hline
				\end{tabular}
				\caption{Derivative-based global sensitivity metrics of all variables on the viscous airfoil problem.}
				\label{tbl:ViscousActivity}
			\end{table*}
			
\subsection{Example 5: Inviscid Transonic Airfoil optimization}
The last demonstration is the optimization of a transonic airfoil in inviscid flow. A CFD code that solves the Euler equation was employed to solve this problem. The design condition for the optimization is $M=0.73$ and $AoA=2^{0}$. The number of initial samples was 45 and enriched with 10 additional samples. The definition of the PARSEC parameterization and the optimization bounds are shown in Table~\ref{tbl:EulerBounds}.
	
			\begin{table}
				\centering
				\begin{tabular}{cccc} \hline 
					No. &	Variables & Lower bound & Upper bound\\ \hline
					1. & $r_{LE}$  & 0.0065 & 0.0092 \\
					2.&$x_{up}$  &   0.3466 & 0.5198\\
					3.&$y_{up}$  &    0.0503 & 0.0755\\
					4.&$y_{xx_{up}}$  &  -0.5094 & -0.3396\\
					5.& $x_{lo}$  &  0.2894 & 0.4342\\
					6.& $y_{lo}$   & -0.0707 & -0.0471\\
					7.&$y_{xx_{lo}}$ &   0.5655 & 0.8483\\
					8.&$\alpha_{te}$ e & -0.1351 & -0.0901\\
					9. & $\beta_{te}$  & 0.1317 & 0.1975\\
\hline
				\end{tabular}
				\caption{Upper and lower bounds of the variables on inviscid transonic airfoil problem.}
				\label{tbl:EulerBounds}
			\end{table}

\begin{figure}
		\centering
		
		\begin{subfigure}{0.7\columnwidth}
			\includegraphics[width=1\columnwidth]{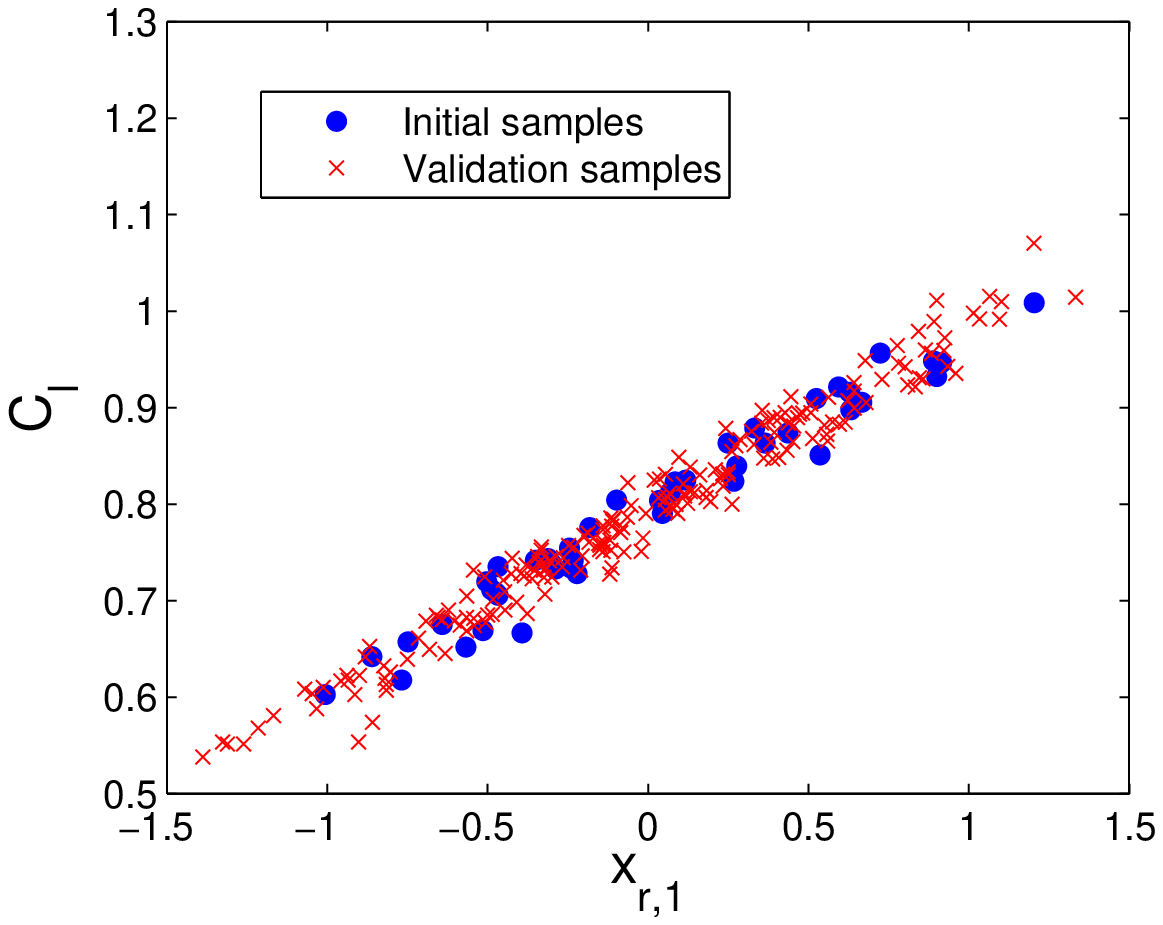}%
			\caption{$C_{l}$}%
			\label{fig:ACTIVESUB_BAR_CL}%
		\end{subfigure}\hfill%
		\begin{subfigure}{.7\columnwidth}
			\includegraphics[width=1\columnwidth]{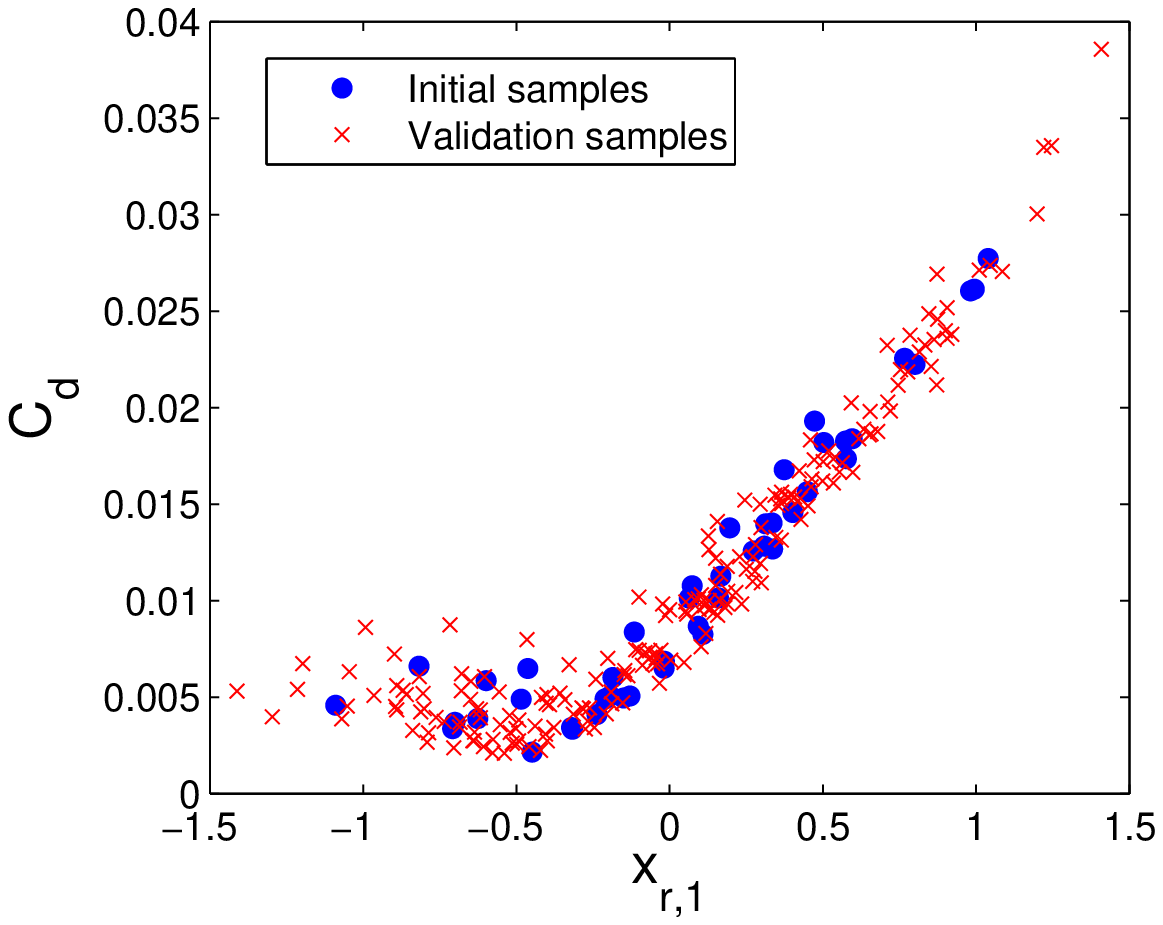}%
			\caption{$C_{d}$}%
			\label{fig:ACTIVESUB_BAR_CD}%
		\end{subfigure}\hfill%
			\begin{subfigure}{.7\columnwidth}
					\includegraphics[width=1\columnwidth]{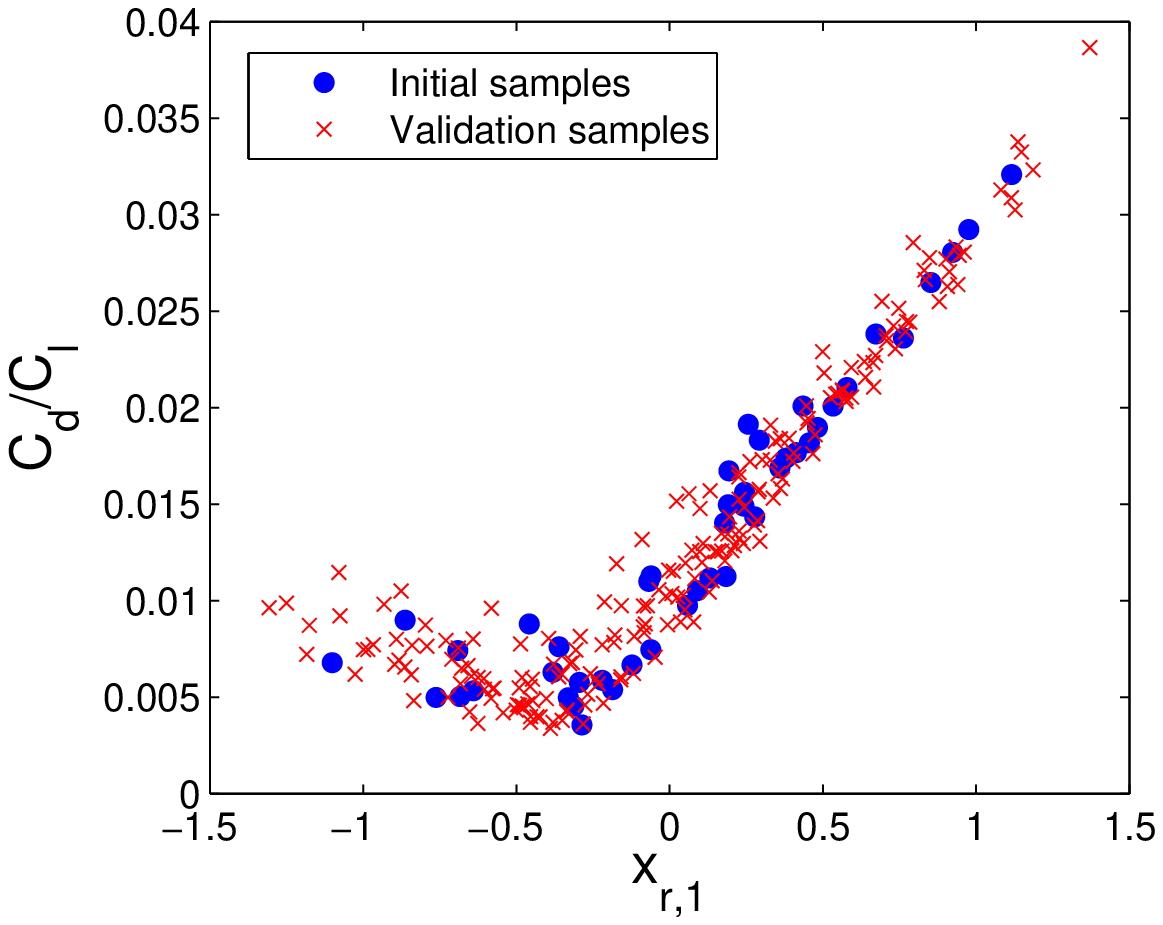}%
					\caption{$C_{d}/C_{l}$}%
					\label{fig:ACTIVESUB_BAR_LD}%
				\end{subfigure}\hfill%
		\caption{Plot in the reduced coordinate for all aerodynamic coefficients on the inviscid airfoil problem.}
		\label{fig:act_inv_airfoil}
	\end{figure}

\begin{figure}
			\centering
			
			\begin{subfigure}{0.7\columnwidth}
				\includegraphics[width=1\columnwidth]{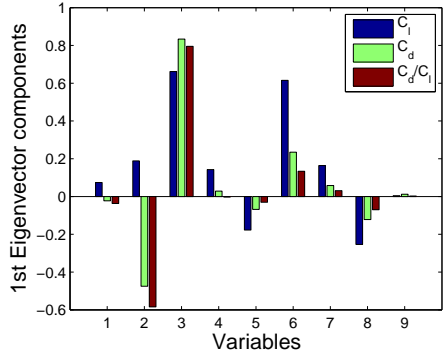}%
				\caption{1st eigenvector components.}%
				\label{fig:ACTIVESUB_AIRFOIL_BAR}%
			\end{subfigure}\hfill%
			\begin{subfigure}{.7\columnwidth}
				\includegraphics[width=1\columnwidth]{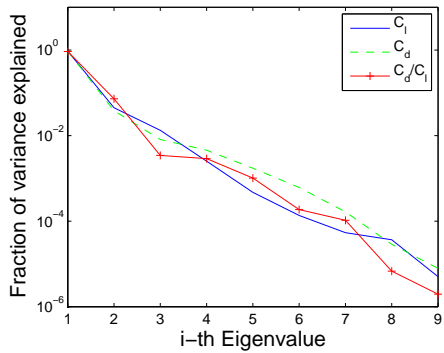}%
				\caption{Eigenvalues decay.}%
				\label{fig:ACTIVESUB_EIGEN_CLCD}%
			\end{subfigure}\hfill%
			\caption{Eigenpairs information on inviscid transonic airfoil problem.}
			\label{fig:braninconvergencae}
		\end{figure}

Our main objective is to minimize $C_{d}/C_{l}$ as the measure of aerodynamic efficiency. However, for the sake of brevity we also show the reduced coordinate plot of the individual objective ($C_{l}$ and $C_{d}$) besides the main objective function as shown in Fig.~\ref{fig:act_inv_airfoil}. Basically, the figure tells us that there is a clear trend of linear behavior for the $C_{l}$ response surface. The response surface of $C_{d}$ and $C_{d}/C_{l}$ are slightly nonlinear and there is a single valley of minimum value near $x_{r,1}=-0.5$ (note that the design space was normalized). Two hundred validation samples were also plotted to further show that the discovered active subspace can really explain the function's variability. Treating $C_{d}/C_{l}$ as the objective function, it is clear from this figure that the problem has a high chance of being a unimodal function, where we can predict the location where the optimum lies. It can also be observed that the one-dimensional active subspace seems to exhibit a quadratic polynomial-like trend. Based on this information one can opt for using a gradient-based local search optimizer on the current solution with minimum function value (such as using adjoint-based gradient), instead of using expensive global search such as evolutionary algorithm. Another choice is to use surrogate-based methods such as ordinary Kriging or universal Kriging. Here, there is an evidence of a quadratic behavior of the response surface, which should be well approximated by a universal Kriging with 2nd order polynomial. Fig.~\ref{fig:ACTIVESUB_AIRFOIL_BAR} depicts the component of the 1st eigenvectors for all aerodynamic coefficients with the plot of eigenvalues decay is shown in Fig.~\ref{fig:ACTIVESUB_EIGEN_CLCD}. The first eigenvalue explained 93.89, 94.68, and 92.03$\%$ of the explained variance for $C_{l}$, $C_{d}$, and $C_{d}/C_{l}$, respectively. On the other hand, the first two eigenvalues explain 98.35, 98.48, and 99.24$\%$ of the explained variance for $C_{l}$, $C_{d}$, and $C_{d}/C_{l}$, respectively. This eigenvalue decay indicates that one or two-dimensional plot in the reduced coordinate is sufficient to explain most of the function's variability. Table~\ref{tbl:EulerActivity} shows that the first and second most important variables for $C_{d}/C_{l}$ are $x_{up}$ and $y_{up}$ with $x_{lo}$ and $\beta_{te}$ are the two least contributive variables. 

\begin{figure}
							\includegraphics[width=0.7\columnwidth]{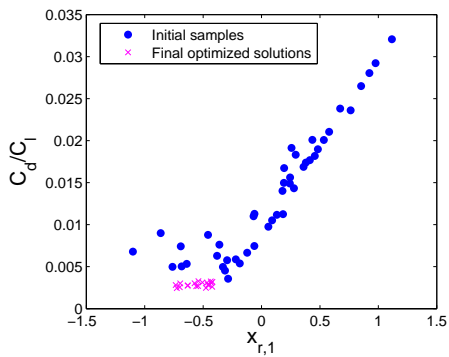}%
							\caption{Plot of one set of initial samples and final optimized solutions in the reduced coordinate.}%
							\label{fig:ACTIVESUB_BAR_CLCDOP}%
						\end{figure}

To test our hypothesis regarding the unimodal nature of the problem, we performed optimization using Kriging-based efficient global optimization (EGO) method~\cite{jones1998efficient} with 20 different sets of initial sampling. The result is then plotted again in the previously computed reduced coordinate and shown in Fig.~\ref{fig:ACTIVESUB_BAR_CLCDOP}. The result shows that the optimized solutions found by EGO in 20 different runs converged to the same valley of the global optimum. This indicates that our unimodal hypothesis for the inviscid transonic airfoil problem was correct. 

							\begin{table*}
										\centering
										\begin{tabular}{cccccccccc} \hline 
											No. & 1 & 2 & 3 & 4 & 5 & 6 & 7 & 8 & 9 \\ \hline
											$\alpha(C_{l})$ & 0.0002 & 0.0013 & 0.0156 & 0.0007 & 0.0025 & 0.0136 & 0.0010 & 0.0023 & 0.0000 \\
											$\alpha(C_{d})\times 10^{4}$ & 0.0022 & 0.5037 & 1.3926 & 0.0055 & 0.0207 & 0.1236 & 0.0094 & 0.0314 & 0.0007 \\
											$\alpha(C_{d}/C_{l})\times 10^{4}$ & 0.0082 & 0.9609 & 1.6111 & 0.0079 & 0.0027 & 0.0533 & 0.0034 & 0.0133 & 0.00001 \\
						\hline
										\end{tabular}
										\caption{Derivative-based global sensitivity metrics of all variables on the inviscid airfoil problem.}
										\label{tbl:EulerActivity}
									\end{table*}
	\section{Conclusion}
In this paper, we demonstrate a framework to perform exploratory analysis of global optimization by analyzing the low-dimensional representation of the problem. The ASM which works by discovering the most active subspaces that capture a large portion of function variability is utilized for this purpose. By rotating the coordinate into the active subspaces, we can take a peek into the behavior and complexity of the objective function. Demonstration on three algebraic and two real-world functions was given to illustrate the usefulness of the framework. The ASM can successfully detect the presence of the single global optimum on the Zakharov function and inviscid transonic problem while it can give a clue regarding the multimodality of the Hartman-6 function. On the four bar trusses problem, Athe SM was able to detect the linear and slightly non-linear characteristics of the first and objective function, respectively. The response surface with non-linear behavior was also successfully observed in the viscous transonic airfoil problem through the ASM. The knowledge obtained from the visualization of low-dimensional representation can then be put to a good use for deciding the type of optimizer or surrogate model to be employed. 

\bibliographystyle{ACM-Reference-Format}
\bibliography{sigproc} 

\end{document}